# Conversation Forests:

# The Key to Fine Tuning Large Language Models for Multi-Turn Medical Conversations is Branching

Thomas Savage MD[1,2,3]
[1] University of Pennsylvania, Perelman School of Medicine, Philadelphia, PA 19104
[2] University of Pennsylvania, Department of Medicine, Philadelphia, PA 19104
[3] University of Pennsylvania, Division of Hospital Medicine, Philadelphia, PA 19104

**Abstract**

Fine-tuning methods such as Direct Preference Optimization (DPO) and Group Relative Policy Optimization (GRPO) have demonstrated success in training large language models (LLMs) for single-turn tasks. However, these methods fall short in multi-turn applications, such as diagnostic patient interviewing, where understanding how early conversational turns influence downstream completions and outcomes is essential. In medicine, a multi-turn perspective is critical for learning diagnostic schemas and better understanding conversation dynamics.  To address this gap, I introduce Savage Conversation Forests (SCF), a reinforcement learning framework that leverages a branched conversation architecture to fine-tune LLMs for multi-turn dialogue. SCF generates multiple possible conversation continuations at each turn, enabling the model to learn how different early responses affect downstream interactions and diagnostic outcomes.  In experiments simulating doctor-patient conversations, SCF with branching outperforms linear training architectures on diagnostic accuracy. I hypothesize that SCF's improvements stem from its ability to provide richer, interdependent training signals across conversation turns. These results suggest that a branched training architecture is an important strategy for fine-tuning LLMs in complex, multi-turn conversational tasks.

**Introduction**

Fine-tuning a Large Language Model (LLM) is the process of adjusting the internal weights of a model to improve performance on a particular task of interest.  In the field of medicine, many researchers have used fine-tuning to inject domain-specific knowledge and improve LLM performance on target tasks such as subspecialty medical expertise[1], patient triage[2], and information retrieval within the electronic health record.[3,4]

  A limitation of popular fine-tuning methods such as Supervised Fine Tuning (SFT), Direct Preference Optimization (DPO)[5], Proximal Policy Optimization (PPO)[6] and Group Relative Policy Optimization (GRPO)[7] is that they are designed for single-turn completions of an isolated prompt. This is a significant limitation in medicine, where many future applications will require multi-turn conversations, with a user communicating across multiple prompt-response exchanges.  The medical interview between a doctor and patient is an especially important example of a potential multi-turn use case.

In a medical interview, physicians ask about a patient's chief complaint and medical history, incrementally gathering information that helps them identify what further questions and information are needed to arrive at the correct diagnosis. Understandably, training an LLM to perform a medical interview with SFT, DPO, PPO or GRPO would be very difficult because they do not allow the model to efficiently learn how one conversational turn influences another.[8] Important interviewing strategies such as disease schemas and the Funnel Technique[9] — initially starting with broad questions and progressively asking more specific and narrow questions — are very difficult to learn from single turn examples alone. Similarly, recognizing common pitfalls such as conversational tangents or dishonest patient responses require a multi-turn perspective that would be difficult to distill with current fine-tuning techniques.

In this manuscript I introduce Savage Conversation Forests (SCF), an adaptation of PPO and GRPO that I have designed for fine-tuning LLMs for multi-turn conversations. My central hypothesis behind SCF is that a branching training architecture facilitates an LLM's ability to learn inter-turn relationships and conversational strategies, thereby enhancing its performance in multi-turn diagnostic interviews.

**Background**

In Reinforcement Learning, two of the most popular algorithms are Proximal Policy Optimization (PPO) and Group Relative Policy Optimization (GRPO).[7]

Proximal Policy Optimization (PPO) was originally proposed by Schulman et al[6] and fine-tunes language models through reward-based learning. PPO involves three models, an actor model that is being trained, a reward model that calculates a scalar reward for actor model completions, and a critic (value) model that estimates a completion's expected reward. To guide learning, PPO compares the reward from the reward model with the expected reward predicted by the value model. This difference — known as the advantage — informs how much and in what direction the policy (actor model) should be updated. PPO also includes regularization mechanisms to constrain policy updates and prevent excessive divergence from the reference model. This is typically done through a probability ratio, clipping, or a Kullback-Leibler (KL) divergence penalty, which help ensure stable training. Figure 1 provides a visual representation of PPO.

Group Relative Policy Optimization (GRPO), introduced by Shao et al[7], is a recent variant of PPO that eliminates the need for a separate critic (value) function. Instead of estimating expected future rewards from a learned critic model, GRPO calculates relative advantages by comparing a group of sampled outputs for the same input. Similar to PPO, each output is scored using the reward model, however, the advantage for each output is computed relative to the mean reward of completions in the group. This group-based normalization enables stable updates without needing value prediction, simplifying the training pipeline while maintaining performance. Figure 1 provides an illustrated representation of GRPO.

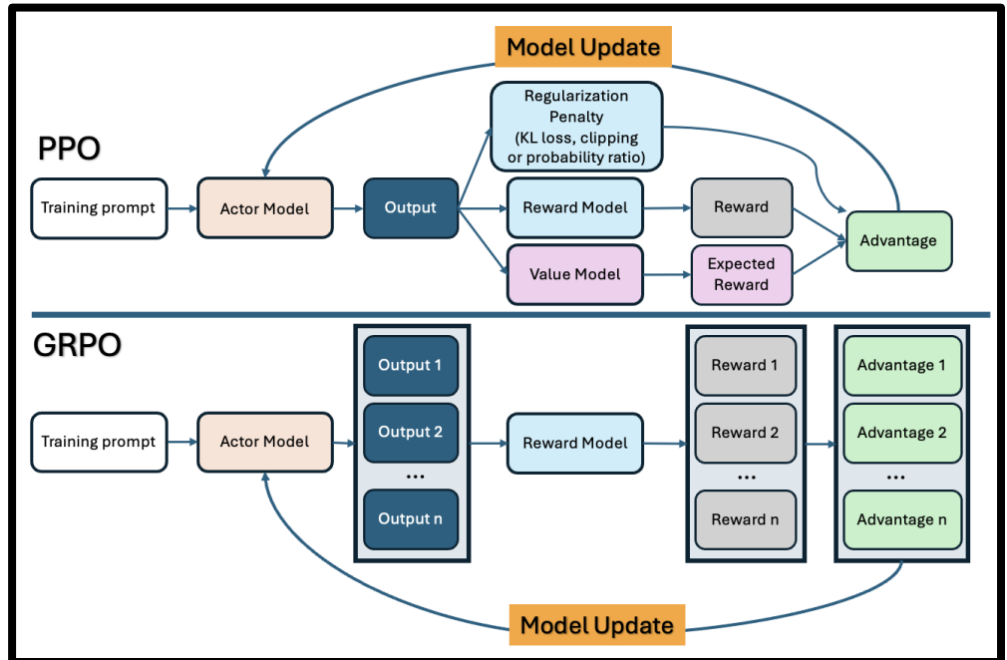

*Figure 1. Visual representation of the Reinforcement Learning methods PPO and GRPO.*

**SCF Architecture**

Savage Conversation Forests (SCF) is a reinforcement learning framework adapted from PPO and GRPO for training large language models (LLMs) for multi-turn conversations. SCF introduces a tree-based architecture that enables connection and learning between conversational turns. While SCF can be applied to a variety of multi-turn applications, this work focuses on its use for training a doctor LLM to take a medical history from a patient.

The base architecture for SCF is GRPO, however instead of grouping across a collection of single turn completions, SCF samples a group of multi-turn conversations. As illustrated in Figure 2, a conversation is simulated between two LLMs: a doctor model (the policy being trained) and a frozen patient model (prompted to simulate a specific chief complaint and diagnosis). Together, they generate a dialogue consisting of alternating turns. After the full conversation is complete, it is passed to a third frozen diagnostician model, which generates a suspected diagnosis. This diagnosis is then compared to a gold standard diagnosis by a frozen grader model, which returns a scalar reward.

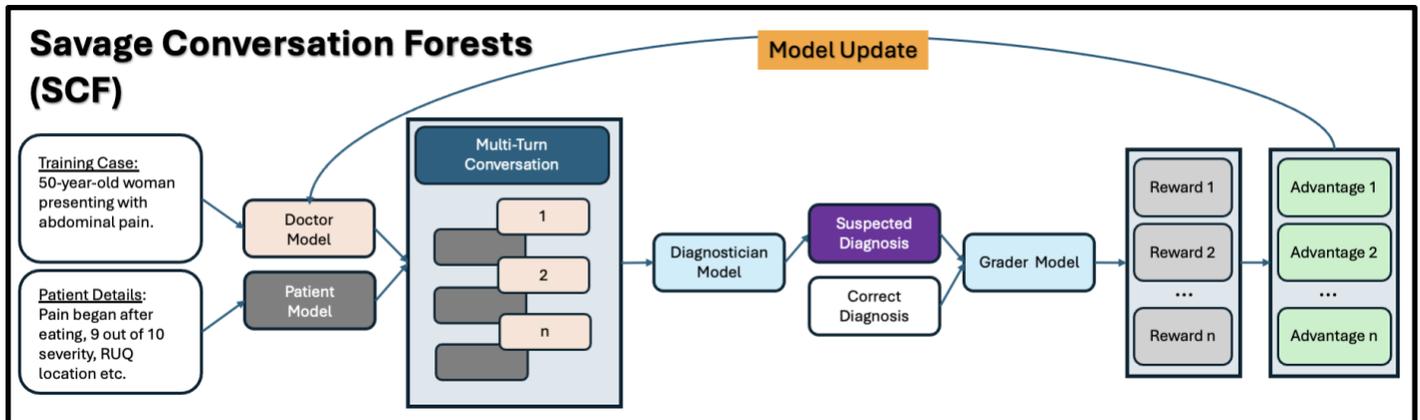

*Figure 2. Visual representation of the SCF method. The multi-turn conversation shown here has been simplified to enhance readability. A comprehensive depiction of SCF's branched multi-turn conversation structure is provided in Figure 3.*

The key innovations of SCF lie in how conversations are structured as well as how rewards are computed and normalized. Instead of sampling linear conversation trajectories, SCF introduces branching at each conversational turn (as shown in Figure 3). Each doctor response spawns multiple possible continuations, resulting in a tree structure: all branches share a common root (the early conversation history) but diverge in their later completions. At the end of each path, or leaf, the grader model provides a scalar reward. Parent branch rewards are in turn calculated by averaging the rewards of their downstream leaves.

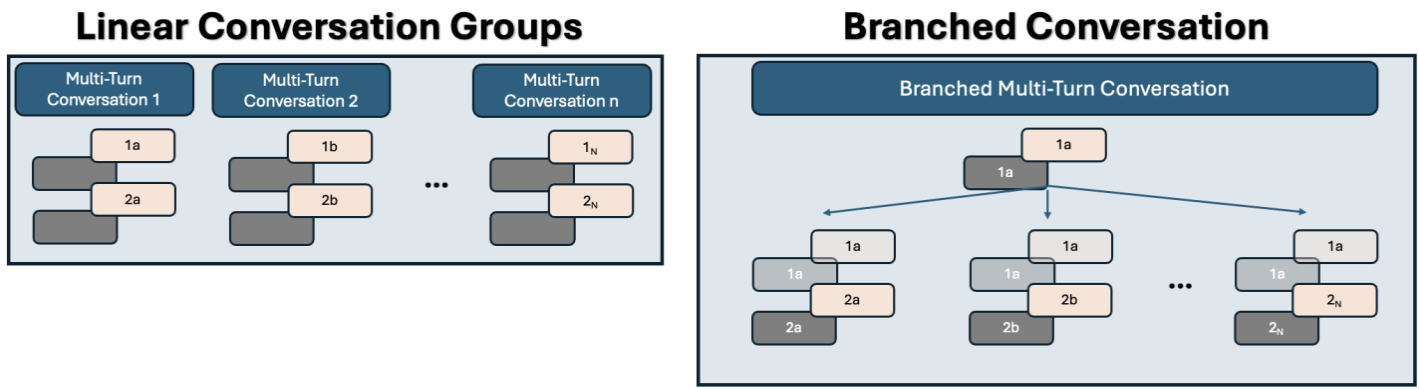

*Figure 3. A comparison of linear conversation structure against the branched conversation structure used in SCF.*

This branching structure creates an imbalance in the number of completions at each conversational level—meaning, there are many more leaves (final turns) than parent branches (early turns). As a result, relative rewards and normalization cannot be uniformly calculated across the group. To address this, branched SCF uses sibling-relative reward calculation: each leaf completion reward is compared only to its sibling leaves (those that share the same immediate parent branch) to compute its relative reward. Then, SCF performs depth-wise normalization, where completions at each conversational depth (e.g., parent or leaf) are normalized only against others at the same level. This ensures that parent branches are only compared to other parents and leaves to other leaves. By combining sibling-relative rewards with depth-aware normalization, SCF maintains both contextual fairness and signal stability. Sibling-relative reward and depth-wise normalization are illustrated below in Figure 4.

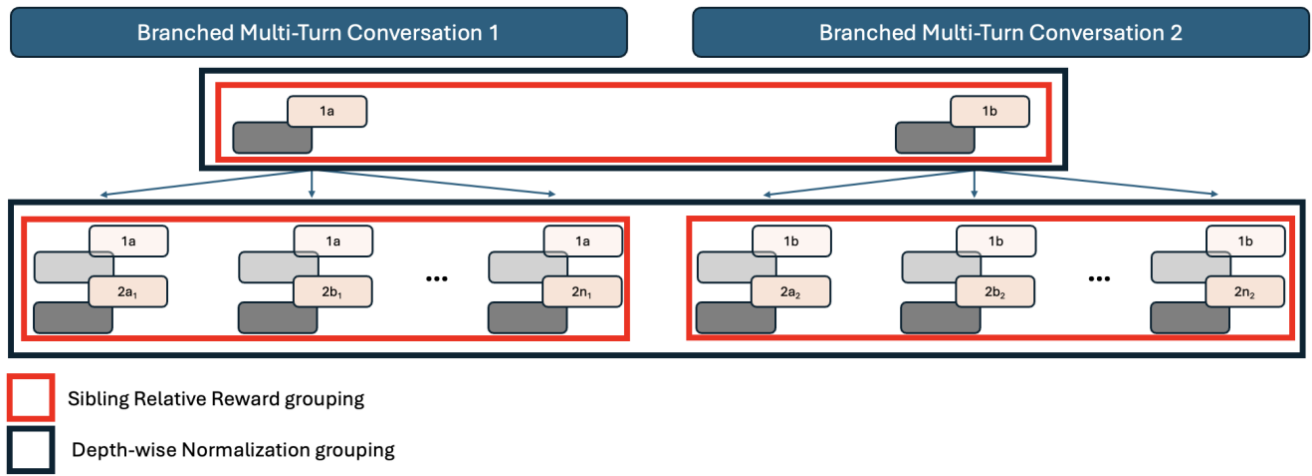

*Figure 4. In SCF the branched conversation structure requires a non-uniform normalization mechanism. The figure above demonstrates the sibling relative reward and depth-wise normalization groupings used in SCF.*

**Related Work - Branched Architecture in Reinforcement Learning**

In the broader reinforcement learning (RL) literature, outside the context of large language models, several studies have explored branching architectures. Notably, Scavuzzo et al[10] and Du et al[11] employed branching structures within Markov Decision Process (MDP) frameworks to solve complex, stepwise problems. Scavuzzo et al focused on optimizing Mixed Integer Linear Programs, while Du et al deployed branching to solve hierarchical recommendation systems. Both works highlight the effectiveness of tree-structured decision processes in domains where a broad problem must be systematically solved in small stepwise increments. This is conceptually analogous to the multi-turn structure of medical interviews, where information must be gathered incrementally through a sequence of questions and answers.

In the context of LLM fine-tuning for multi-turn conversations, one of the most relevant prior works is SWEET-RL, recently proposed by Zhou et al.[12] SWEET-RL fine-tunes language models for multi-turn conversations by applying a modified Direct Preference Optimization (DPO) step at each conversational turn. At every turn, SWEET-RL samples multiple completions that are scored using a trained preference model. The preference order is used to train the model using DPO and a top-ranked response is selected to advance the conversation. Although this sampling-and-ranking approach allows for local optimization at each turn, SWEET-RL advances the dialogue using a linear rather than a branching trajectory. As a result, it does not foster the contextual diversity needed to explore how different questions or statements might alter downstream conversation paths or diagnosis accuracy. In contrast, SCF generates and retains full sets of sibling completions at each turn, allowing the model to observe and learn from how multiple alternative completions shape the conversation's future progression.

To my knowledge, SCF is the first method to combine explicit tree architecture, sibling-relative reward calculation, and depth-aware normalization within a reinforcement learning framework for multi-turn LLM training. SCF's structure enables the model to learn how conversational turns relate to one another within the broader context of a diagnostic interview—a signal that existing methods such as PPO, GRPO, or SWEET-RL are not designed to capture.

**Methods**

To evaluate the effectiveness of SCF, we compared two variants of the method: one using a linear conversation structure (i.e., a single branch per turn) and one using a branched tree structure (i.e., four branches per turn). Both variants were tested on a modified version of the MedQA dataset[13], adapted to support multi-turn doctor-patient interactions. We conducted experiments using two popular open-source LLMs—Llama-3.1-8B-Instruct[14] and Ministral-8B-Instruct[15] to assess the generalizability of SCF across model architectures.

Medical Conversation Dataset

To evaluate SCF in the context of multi-turn medical interviews, we constructed a custom dataset derived from the publicly available MedQA dataset[13], which contains U.S. medical board examination questions describing clinical cases. To adapt this dataset for dialogue-based training, we used GPT-4o to programmatically extract structured components from each MedQA question-answer pair, including:

- A concise, one-line case introduction (e.g., patient age, gender, and chief complaint)
- A detailed list of relevant clinical information, such as medical history and presenting symptoms
- The final diagnosis as specified (or inferred) by the original MedQA answer key

This transformed dataset enables the simulation of realistic patient responses during training and provides gold-standard diagnostic labels for reward evaluation. The code used to generate the dataset is available in Appendix I, and the dataset itself is provided in Appendix II.

Code and Computing Specifications

All code for implementing Conversation Forests, including both the branched and linear variants, is publicly available at github.com/tsavage6001/SCF. Model training was conducted using a high-performance computing cluster equipped with three A100 PCIe GPUs.

The role of the diagnostician and grader models was performed by independent instances of GPT-4o[16], accessed between May 15th – July 1st, 2025.

### Hyperparameters

Each training run used a conversational depth of two turns, meaning that every simulated dialogue consisted of two doctor questions and two corresponding patient responses. To ensure parity in training signal across SCF variants, we matched their total number of generated completions per case:

- In the branched SCF variant (branching of 4), we generated 4 conversation trees per case.
- In the linear SCF variant (branching of 1), we generated 10 conversation trees per case.

This ensured that each variant produced 20 total completions per training case, enabling a fair comparison between the two architectures.

Finally, all models were trained with a learning rate of 2e-7, AdamW optimizer, and a maximum generation length of 20 tokens per response. Full precision was used for all model training. Training cases with insufficient reward variability (e.g. all responses for a case were given the same reward) were skipped and did not participate in training.

### Model Evaluation Metrics

Each trained model was evaluated on a held-out test subset of the conversation dataset. Performance was assessed using the same diagnostician and grader models employed during training. For each test conversation, the model generated a linear conversation transcript and the diagnostician model generated a corresponding predicted diagnosis. This prediction was then scored by the grader model against the ground-truth diagnosis label. The grader model assigned one of three possible values:

- 1.0 for an exact or clinically equivalent diagnosis,
- 0.5 for a partially correct or related diagnosis,
- 0.0 for an incorrect diagnosis.

Model performance is reported as the percentage of total possible points earned across all test cases.

### Response Evaluation Metrics

To assess differences in response-level model output resulting from linear versus branched training strategies, I conducted a detailed analysis on a randomly selected set of 100 test conversations from the Llama3 test set. Specifically, the evaluation focused on the broadness (or open-endedness) of the questions posed by each doctor model.

The broadness of each doctor-generated question was evaluated on a 5-point Likert scale, ranging from 1 (most broad/open-ended) to 5 (most narrow/closed). The criteria for this scale are provided in Appendix III. Each question was independently reviewed and scored by three medical providers, and the average score was calculated for each response. The mean broadness scores of the linearly trained model were then compared to those of the branched model using a paired t-test, with a significance threshold of α = 0.05.

**Results**

Our results demonstrate that SCF with branching outperforms both the linear SCF variant and the base model. When trained with the branched SCF architecture, a fine-tuned Llama-3.1-8B-Instruct model achieved a diagnostic accuracy of 49.2%, while a fine-tuned Mistral-8B-Instruct model achieved 48.8%. In contrast, the linear SCF variant yielded lower or roughly equivalent performance relative to the base models: 45.4% for Llama and 36.2% for Mistral. The base models, without any SCF fine-tuning, scored 45.1% and 45.5%, respectively.

|  | Base Model | Linear SCF | SCF with Branching |
| --- | --- | --- | --- |
| Llama-3.1-8B-Instruct | 45.1% | 45.4% | 49.2% |
| Ministral-8B-Instruct | 45.5% | 36.2% | 48.8% |

*Table 1. Results of our investigation comparing linear SCF and SCF with branching in performing multi-turn medical interviews. Model performance is reported as the percentage of total possible points earned by the model across all test cases.*

The training dynamics for the branched and linear SCF variants using Llama are shown in Figure 5. The figure illustrates that branched SCF consistently achieved higher mean rewards throughout training.

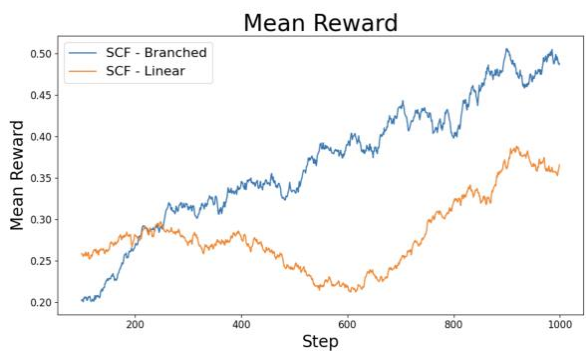

*Figure 5. Plot of the mean reward for Llama when training with branched (blue) and linear (orange) SCF.*

I evaluated the robustness of my findings by testing two enhancements to the linear SCF variant: pre-fine-tuning with supervised learning (SFT) and incorporating a KL-divergence penalty during training. Neither strategy led to improvements in diagnostic accuracy. Across all experiments, the branched SCF architecture consistently outperformed the linear variant.

To better understand how training structure influenced model behavior, I analyzed both the broadness and lexical diversity of the questions generated by Llama3 models trained with linear versus branched SCF. Table 2 summarizes the broadness of questions asked by both models, scored on a 5-point Likert scale where lower scores correspond to more open-ended questions. The branched model consistently produced questions that required more detailed answers.

|                    | Linear SCF Llama3 | Branched SCF Llama3 | p-value |
|--------------------|-------------------|---------------------|---------|
| Question Broadness | 3.1               | 2.7                 | 0.04    |

*Table 2. Broadness and degree of open-endedness for questions posed by Llama3 trained with Linear SCF and Branched SCF. Lower Likert scores correspond with broader and open-ended questions.*

Table 3 then presents the most common 5-grams for each model to provide concrete examples of these stylistic differences. The combination of expert scoring and n-gram examples demonstrate that the branched SCF model produced questions that were more open-ended.

| Top 5-grams of Linear SCF Llama3 | Top 5-grams of Branched SCF Llama3 |
|---|---|
| 1. 'what is the duration of' (12) | 1. 'have you noticed any other' (9) |
| 2. 'is the duration of these' (4) | 2. 'what is the duration of' (8) |
| 3. 'is the duration of your' (4) | 3. 'you noticed any other symptoms' (8) |
| 4. 'duration of your abdominal pain' (3) | 4. 'what was the onset of' (7) |
| 5. 'have you noticed any other' (3) | 5. 'is the duration of symptoms' (6) |
| 6. 'the duration of these symptoms' (3) | 6. 'the duration of your symptoms' (4) |
| 7. 'the duration of your abdominal' (3) | 7. 'was the onset of symptoms' (4) |
| 8. 'what the duration of your' (3) | 8. 'medications are you currently taking' (3) |
| 9. 'you noticed any other symptoms' (3) | 9. 'what been the duration of' (3) |
| 10. 'do you have any history' (2) | 10. 'what is your medical history' (3) |

*Table 3. Top ten 5-grams for Llama3 trained with Linear SCF (left) versus Branched SCF (right). The numbers in parentheses are the number of instances of each 5-gram in the corresponding dataset.*

**Discussion**

The findings of my study demonstrate that incorporating a branched training structure through SCF significantly improves model performance on multi-turn medical interviews compared to a linear architecture. As shown in Figure 5, the branching design enables earlier and more robust recognition of training signals. I suspect this improvement stems from the SCF's ability to receive richer feedback about how early conversational turns influence downstream completions and, ultimately, diagnostic accuracy. By capturing interdependencies between turns, the branched SCF architecture better supports the training of effective multi-turn conversational agents.

While the absolute improvement in diagnostic accuracy achieved by branched SCF may appear modest (3-4%), this is likely due to the limited conversational depth of my experiment, which included only two turns per dialogue. With such a short exchange, there are relatively few opportunities for the model to demonstrate and benefit from deeper inter-turn reasoning. I expect that increasing the number of conversation turns would further amplify the performance gains of branched SCF, as the benefits of modeling long-range dependencies would become more pronounced.

To better understand how training structure influences model behavior, I compared the types of questions generated by models trained with linear versus branched SCF. The branched SCF model consistently produced broader and more open-ended questions (Table 2), prompting richer patient responses. For example, it favored phrasings like "have you noticed any other" [symptoms], which encourage the patient model to provide longer, more detailed answers. In contrast, the linear SCF model tended to ask narrower, more specific questions such as "what is the duration of," which typically elicit shorter responses (Table 3). These differences suggest that the branched SCF architecture allows the model to learn that open-ended questions can yield more informative responses—an insight that does not emerge as clearly from a linear training structure.

The primary limitation of SCF lies in its computational complexity. The branching design results in an exponential increase in the number of conversation paths as more turns are added, leading to significant resource demands. As a result, this study was restricted to evaluating SCF on short conversations and on models with up to eight billion parameters. Evaluating SCF on longer conversations and larger models represents an important direction for future work.

Future research should explore whether branched SCF enables models to learn more advanced conversational dynamics relevant to clinical practice. One promising direction is investigating whether models can autonomously learn disease schemas to support systematic diagnostic reasoning. Another important area is examining whether branched SCF helps models better identify common conversational pitfalls in medicine, such as recognizing when a patient response veers off-topic or when vague answers require targeted follow-up. Such nuanced behaviors are difficult to capture in short dialogues and likely require greater conversational depth than was feasible given the resource constraints of this study. I encourage others in the field to further test SCF in more complex settings to better understand how branching architectures can support the emergence of sophisticated, medically meaningful dialogue strategies that are difficult to learn through single-turn training alone.

Finally, I believe this investigation has implications that extend beyond the domain of medicine. The results strongly suggest that a branched training architecture offers meaningful advantages when fine-tuning LLMs for multi-turn tasks. I encourage researchers in other fields that rely on multi-turn interactions, such as education, law, and engineering[8], to explore the application of SCF within their respective domains. Broader testing across diverse use cases will be essential to fully validate and refine this training methodology.

**Conclusion**

The branched training architecture of SCF results in better model performance for multi-turn medical interviewing compared to a linear architecture. Branching enables the model to better learn how early completions affect the conversation's direction, encourages more open-ended questions, and ultimately improves diagnostic accuracy.

**Appendix I**

The file used to adapt the MedQA question-answer dataset to a dataset for medical conversations is attached as the file: create_multi_turn_convo_dataset.ipynb.

**Appendix II**

Modified MedQA dataset for multi-turn medical interviews.

**Appendix III**

Question broadness Likert scale criteria.

| Score | Description |
|---|---|
| 1 | Requires multiple sentences to answer |
| 2 | Requires a sentence to answer |
| 3 | Requires a few words answer |

| 4 | Requires a single word answer |
| 5 | Requires a single yes-no answer |